# Integrating LLMs in Gamified Systems

CARLOS J. COSTA
ISEG – Lisbon School of Economics and Management
University of Lisbon
cjcosta@iseg.ulisboa.pt
Lisboa, PORTUGAL

*Abstract:* - In this work, a thorough mathematical framework for incorporating Large Language Models (LLMs) into gamified systems is presented with an emphasis on improving task dynamics, user engagement, and reward systems. Personalized feedback, adaptive learning, and dynamic content creation are all made possible by integrating LLMs and are crucial for improving user engagement and system performance. A simulated environment tests the framework's adaptability and demonstrates its potential for real-world applications in various industries, including business, healthcare, and education. The findings demonstrate how LLMs can offer customized experiences that raise system effectiveness and user retention. This study also examines the difficulties this framework aims to solve, highlighting its importance in maximizing involvement and encouraging sustained behavioral change in a range of sectors.

## 1 Introduction

Gamification has become a potent instrument for raising user motivation and engagement in a variety of industries, including business, healthcare, and education. [3] [13] By integrating game features like challenges, rewards, and feedback in non-gaming contexts, gamified systems have enhanced user engagement, learning outcomes, and behavioral change. It is possible to further improve these systems by enabling adaptive learning, personalized feedback, and dynamic content generation with the rise of cutting-edge technologies like Large Language Models (LLMs). [15]

Although gamification and LLMs have been examined separately, little is known about how they can be combined into a single framework, particularly in mathematical modeling. Not considering the potential of LLMs and the growing interest in gamification, not much has been done to systematically combine these technologies into a coherent framework. The study reported in this paper examines the framework for integrating LLMs into gamified systems to improve task dynamics, user engagement, and system performance. The study specifically aims to comprehend how LLMs can impact behavior learning and engagement in various real-world applications.

Creating a framework that formally incorporates LLMs into gamified systems is the primary goal of this research. The framework will incorporate dynamic and personalized capabilities and concentrate on user engagement, task adaptation, and reward mechanisms. In order to clarify the potential advantages of LLM-enhanced gamification, the study also studies the real-world applications of this framework in sectors like retail, healthcare, and education. The purpose of developing a systematic method for incorporating LLMs into gamified systems is to close the gap between theoretical models and practical applications.

The basis of the study's methodological approach includes application exploration, simulation design, and mathematical framework development.

The research reported here is relevant because it presents a thorough mathematical framework for incorporating Large Language Models (LLMs) within gamified systems.

Although earlier research has examined LLMs and gamification independently, formalized mathematical models that clearly articulate user engagement dynamics, task adaptation, and reward optimization have been scarce. This study introduces an innovative mathematical model, validates it through simulation, and offers a framework applicable across various industries.

Here, we present differential equations that quantitatively characterize reward mechanisms, task adaptation, and user engagement. This study shows the framework's efficacy across several domains by implementing and testing it in a simulated environment in contrast to earlier conceptual studies. The method described in this paper is very flexible and can be used in retail healthcare and education settings. By offering a systematic data-driven method for gamified experience optimization with LLMs, this research closes a significant gap and significantly contributes to academia and business..

## 2 Problem Formulation

Gamification and artificial intelligence (AI) integration have gained significant attention across multiple fields.

### 2.1 Gamification in Education and Healthcare

Gamification has been widely used in education to improve learning outcomes and student engagement. Many studies show how game-based approaches can promote student motivation and active learning [1][7][9][12]. Gamification improves engagement and learning outcomes in education [1] [2]. The same occurred with gamified healthcare systems. Using challenges and rewards motivates users and promotes behavior change [2]. Gamified healthcare systems also have demonstrated potential for enhancing patient compliance and promoting healthy behaviors [14] [25] [26] [27].

### 2.2 Role of LLMs in Adaptive Learning

The LLMs' ability to produce context-aware and personalized content has shown much promise [3][16][17]. For example, Sun and colleagues researched how LLMs could improve adaptive learning systems by customizing content based on user preferences and performance [3]. Additionally, LLMs dynamically modify task difficulty and enable real-time feedback, assuring constant levels of engagement [4]. LLMs are perfect for adaptive learning settings because of their well-known capacity to produce customized content [4]. The user experience is improved by these models' ability to modify tasks and offer real-time feedback dynamically. [15] [22], [23], [24].

### 2.3 Modeling Gamification

The theory suggests that to be effective, the gamification needs to find a balance between user engagement and the difficulty of tasks [5]. Gamification has evolved so much in the last few years, which may be verified by the increasing number of frameworks proposed by researchers or practitioners [7]. Some authors focus on the significance of reward structures in maintaining user engagement, pointing out the necessity for flexible systems [6]. Other authors studied models aiming to enhance gamified systems [10], including aspects like task progression and reward allocation [5],[6].

## 3 Method

The methodological approach used in the research reported in this paper is deductive, involving developing a mathematical model. Then, I designed a simulation to test and explore implications. Finally, some real-world scenarios were explored.

### 3.1 Mathematical Framework Development

A model for user engagement task adaptation and reward optimization was developed. The primary equations encompass user engagement dynamics, task adaptation, and reward optimization.

- User Engagement Dynamics:

$$dE/dt = \alpha R(t) - \beta D(t)$$

- Task Adaptation:

$$T_i(t+1) = T_i(t) + \gamma (U(t) - S_i(t))$$

- Reward Optimization:

$$R(a_t) = w_1 G(a_t) - w_2 C(a_t)$$

These equations are the theoretical basis for combining LLMs into gamified environments.

### 2. Simulation Design

A simulated environment was developed to evaluate the framework. The simulation consisted of 3 steps. First, user profiles featuring variable engagement rates were defined. •Progression of task difficulty contingent upon user performance. Then, a reinforcement learning algorithm was designed to enhance user actions.

### 3. Application Exploration

The framework was analyzed for its applicability across industries. Specific use cases were outlined to demonstrate adaptability and impact.

## 4 Results

### 4.1 Mathematical Framework Development

The framework previously described is developed by combining the analysis of user behavior (applies behavioral modeling), the dynamics of tasks (using dynamic systems theory), and optimization methodologies (reinforcement learning concepts).

#### 4.1.1 User Engagement Dynamics

A differential equation is employed to characterize the rate of change in user engagement, $E(t)$, as it is affected by both reward mechanisms and factors contributing to disengagement.

$$dE/d_t = \alpha R(t) - \beta D(t)$$

Where:
- R(t) is the Reward rate and represents how the system motivates the user.
- D(t) is the disengagement rate, describing how external or internal factors cause users to lose interest.
- α and β are coefficients that adjust the impact of R(t) and D(t) on engagement.

This model explains the user retention prediction and identifies optimal reward structures for sustained engagement.

### 4.1.2 Task Adaptation
In gamified systems, it is essential for tasks to adapt to user performance. This approach employs iterative adjustment models to modify task difficulty, $T_i(t)$, by taking into account user performance U(t) and the success rate of tasks $S_i(t)$.

$$T_i(t+1) = T_i(t) + \gamma(U(t) - S_i(t))$$

Where:
- U(t) Measures the user's ability, such as success rate or completion time.
- $S_i(t)$ tracks the system's measure of success for the task.
- γ a parameter controlling how quickly tasks adapt.

This approach ensures that tasks remain appropriately challenging, avoiding frustration or boredom.

### 4.1.3 Reward Optimization
The methodology employs decision-making models to maximize rewards by applying temporal motivation theory. The reward R(at) for a specific action at is determined as follows:

$$R(a_t) = w_1 G(a_t) - w_2 C(a_t)$$

Where:

- G(at) is the long-term gain from the action, such as learning a skill or achieving a goal.
- C(at), The immediate cost, such as time or effort required.
- w1,w2 >0, Weights that balance the importance of long-term benefits vs. short-term costs.

This model provides a mathematical basis for determining the best actions to encourage desired user behaviors.

### 4.1.4 Reinforcement Learning
Reinforcement learning may also be used to refine user interaction. Specifically, it uses Q-learning, which updates the quality (Q) of an action $a_t$ taken in a state $s_t$:

$$Q(s_t, a_t) = r_t + \delta \max Q(s_{t+1},)$$

Where:
- $Q(s_t, a_t)$ measures the value of acting at in state st.
- $r_t$ is the immediate reward received after performing $a_t$.
- δ is the discount factor, weighing future rewards.

This model helps improve system responses to increase user satisfaction and engagement.

### 4.2. Simulation Outcomes
Figure 1 shows the impact of changing α, β, and γ on user engagement and task difficulty.

The chart indicates that an increase in α (reward coefficient) results in a more rapid user engagement growth. In addition, the data shows that a higher β (disengagement coefficient) leads to a more noticeable decline in user retention. Besides, the chart illustrates the influence of γ (task adaptation rate) on the fluctuations in task difficulty. This picture allows an intuitive comprehension of how parameter modifications influence the system's overall behavior.

The simulation confirmed that the framework proposed in this paper contributes to effectively maintaining user engagement, maximizing rewards, and adjusting tasks in real-time. Performance indicators demonstrated enhanced user retention and increased success rates when generative AI is incorporated into gamified environments.

The findings from the simulation offer a meaningful understanding of the relationship between user engagement, task adaptability, and reward systems in a gamified context augmented by using generative artificial intelligence. These findings support the validity of the proposed framework.

representing time intervals and the y-axis indicating difficulty levels. The adaptation of task difficulty is influenced by user performance (U(t)) and the desired success rate (Si(t)) following the update rule established by the framework:
- Dynamic Adjustments
- Stable Difficulty Levels

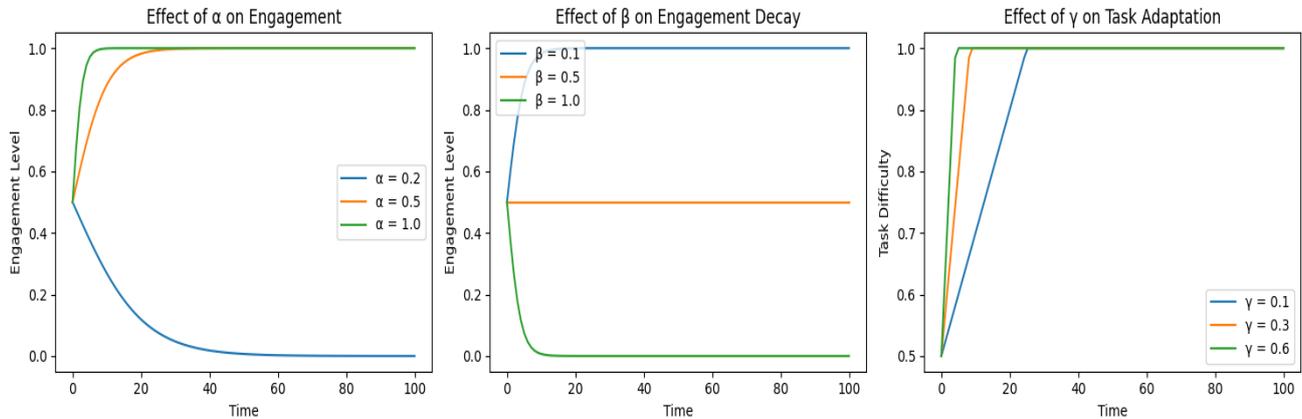

Figure 1 – Effect of α, β, and γ on user engagement and task difficulty over time

### 4.2.1 User Engagement Over Time

The initial chart in Figure 1 shows the progression of user engagement (E(t)) throughout the simulation period from steps 0 to 100. The horizontal axis indicates the time steps, while the vertical axis shows the user's engagement level, which ranges from 0 to 1. The engagement level exhibits variations over time influenced by the dynamics of rewards (R(t)) and factors contributing to disengagement (D(t)): early stage, intermediate variations, and long-term stabilization.

In the early moments of the simulation, engagement generally increases significantly due to substantial rewards that motivate the user.

As the system evolves and the complexity of tasks changes, user engagement experiences fluctuations that correlate with the user's ability to perform the challenges he/she faces.

Over time, engagement levels generally reach a state of stability or exhibit cyclical variations, indicating the system's capacity to adaptively modulate the challenges associated with sustaining user interest and motivation, illustrating the efficacy of the LLM-driven framework in regulating engagement by employing a judicious mix of incentives and penalties for disengagement, thereby enabling the system to enhance user retention.

### 4.2.2 Task Difficulty Adaptation

The second chart shows the progression of task difficulty (Ti(t)) as a function of time, with the x-axis

- Insights

Task difficulty is designed to escalate when the user demonstrates proficiency, while it diminishes if performance falls short of the established target. The difficulty level sooner or later stabilizes as the system refines its adaptive capabilities, maintaining an equilibrium that prevents tasks from becoming excessively simplistic or extremely challenging for the user. This mechanism of adaptive difficulty is essential for continuing user engagement. A fast increase or decrease in difficulty can lead to user frustration or a lack of challenge. This adaptability focuses on the fundamental advantage of incorporating large language models into gamified environments: real-time adjustments to difficulty based on user performance to ensure an optimal experience.

### 4.2.3 Interrelationship Between Engagement and Difficulty

From game practice and studies expressed in flow theory [31], engagement is expected to diminish when the difficulty of a task exceeds the user's capabilities. Consequently, users may feel crushed and disengaged if the challenge is too difficult. On the other hand, when tasks are excessively simplistic, users may experience a decline in interest due to insufficient challenge, which can also lead to reduced engagement.

The most effective engagement occurs when task difficulty is continuously adjusted to the user performance, achieving an equilibrium between challenge and reward.

These insights emphasize the dynamically calibrated task difficulty needed to maintain user engagement over long periods of time.

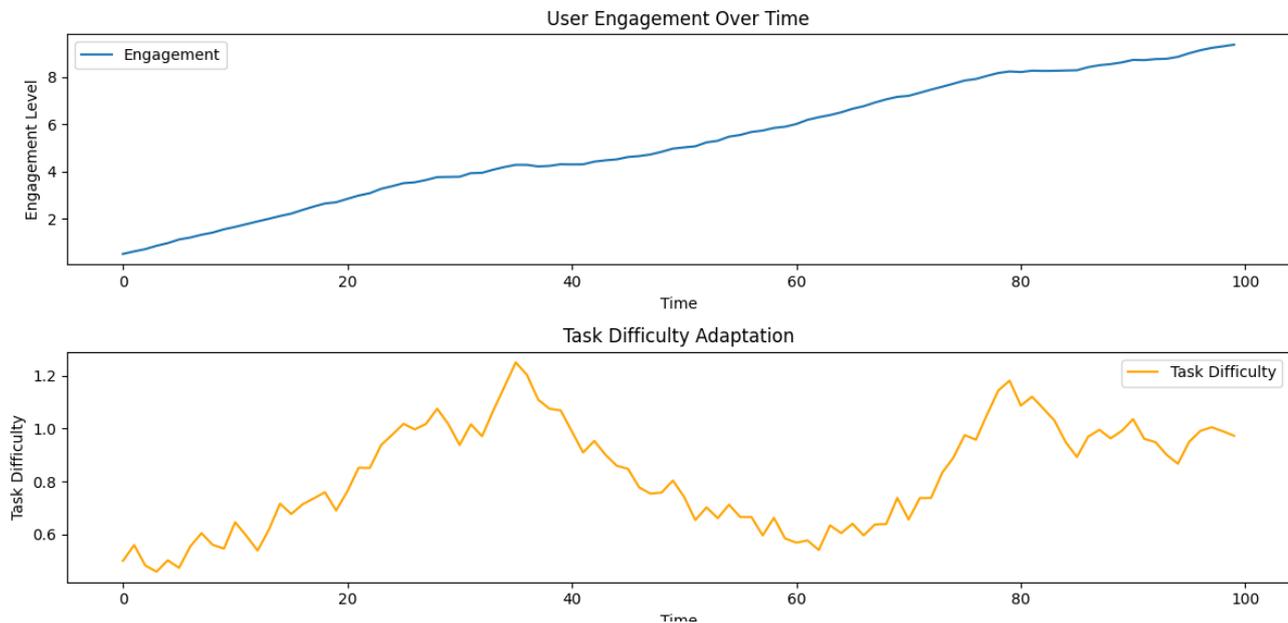

Figure 2 – User engagement over time and Task difficulty adaptation

### 4.2.4 Sensitivity to Parameter Settings

The results of the simulation further illustrate the system's sensitivity to critical parameters, which include α (reward scaling), β (disengagement scaling), and γ (adaptation rate).

Reward scaling (α) affects the rate at which engagement responds to rewards. A higher α value results in an immediate increase in engagement upon introducing rewards.

Disengagement scaling (β) influences disengagement in the system. A higher β value leads to a speedier decline in engagement when disengagement factors are pronounced.

Adaptation rate (γ) regulates the speed at which task difficulty is modified in response to user performance. A higher γ enables quicker adjustments, although it may also result in sudden changes in task difficulty.

By manipulating these parameters, the system can be optimized to deliver a customized experience, ensuring the simulation's applicability across various contexts.

### 4.2.5 Summary of Findings

The simulation confirms the efficacy of the proposed framework. It illustrates its capacity to improve user engagement and tailor task difficulty within gamified settings. The system can dynamically modify task challenges and rewards by incorporating LLMs, maintaining user interest and motivation.

The following section explores the implications of these results for multiple industries, emphasizing the practical uses of this framework in fields such as education, healthcare, and entertainment.

## 4.3 Applications Across Industries

Table 1 presents a comprehensive overview of the potential applications of the framework across various industries, outlining specific objectives, implementation methodologies, and anticipated outcomes. This table shows how integrating artificial intelligence into gamified systems can enhance user engagement, task dynamics, and overall system efficacy across different sectors. It emphasizes the goals and strategies for implementation and the expected benefits of employing artificial intelligence in many industries or other human activities.

Gamification has already made significant progress in education and produced positive results. However, incorporating generative artificial intelligence still offers opportunities for customizing educational experiences. For example, adaptive learning tasks can be implemented using LLMs, delivering real-time feedback and adjusting the difficulty level answering to a student's performance. For example, a student encounters difficulties with mathematics. The system could generate additional practice problems, simplify explanations, or provide alternative learning resources designed according to the student's needs. This customized approach may improve educational outcomes and reduce abandon rates, as students are more likely to engage with material that aligns with their current understanding and learning pace.

In healthcare, gamification and serious games have produced varied outcomes. For instance, AI-generated health can be customized to align with an individual's fitness level, such as proposing a walking experience for beginners and experienced runners. The LLM can provide real-time guidance and feedback, modifying the challenge's intensity based on the user's progression. Patients who receive adapted advice and challenges are more persuaded to remain motivated and adhere to health regimens. Obviously, the result will be an improved long-term health outcome.

Table 1: Applications of the LLM-Integrated Gamified Framework

| Industry | Objective | Implementation | Expected Impact |
|---|---|---|---|
| Education | Enhance student engagement | Adaptive learning tasks using LLMs for real-time feedback and content personalization | Improved learning outcomes and retention rates |
| Healthcare | Motivate healthy behaviors | Personalized health challenges and fitness tracking with AI-generated advice | Better patient compliance and healthier lifestyles |
| Retail | Increase customer loyalty | AI-driven rewards programs tailored to shopping behavior and preferences | Higher customer retention and increased sales |
| Corporate Training | Improve employee skills | Scenario-based training simulations with real-time feedback and dynamic difficulty adjustment | Accelerated skill acquisition and employee satisfaction |
| Entertainment | Create immersive experiences | Interactive storytelling powered by LLMs that adapt narratives to user choices | Enhanced user engagement and satisfaction |

While gamification has been a research subject in several business areas, LLMs present an opportunity to enhance customer loyalty through personalized rewards programs. For example, an online retailer could offer customized discounts or exclusive deals based on a customer's past shopping behavior. The LLM can monitor a customer's preferences and purchase history to deliver adapted recommendations and rewards that resonate with individual tastes. Such a high degree of personalization can foster greater customer retention, as individuals feel valued and are more likely to return for future purchases.

LLMs can significantly improve employee training in many companies. An area of application could be facilitating scenario-based simulations. For example, a customer service representative could participate in interactive role-playing scenarios driven by LLMs. These simulations would adapt in real-time, increasing complexity as the employee's skills develop. Feedback from the LLM could provide guidance and suggest improvements for the employee's responses in various situations. This adaptive training approach accelerates skill development and may contribute to employee satisfaction as the training experience becomes more relevant and engaging. Consequently, this approach can amplify the effectiveness of gamification initiatives.

LLMs have the potential to change interactive storytelling by customizing experiences to user decisions within the context of entertainment. For instance, the narrative can change in a video game context in response to the player's choices. However, if this approach is improved using generative artificial intelligence tools, like language models, it can create new dialogues or develop plots that reflect the player's actions. This results in an exclusively personalized experience, where the storyline is fluid and evolves according to the player's engagement with the game. Subsequently, users are more likely to remain engaged and satisfied, perceiving their decisions as directly impacting the narrative's progression.

Additionally, generative artificial intelligence tools play a critical role in enhancing gamified systems through their ability to facilitate real-time adjustments, provide tailored feedback, and generate dynamic content. In education, for example, a generative AI-based system can evaluate student inputs and produce adapted practice exercises, explanations, or hints, modifying the difficulty based on prior interactions. In healthcare, gamification driven by artificial intelligence can foster patient engagement by adjusting health-related challenges in real time, delivering AI-generated motivational messages, and offering suggestions that align with user progress. In retail, customer engagement can be elevated through AI-enhanced, gamified loyalty programs, where LLMs analyze consumer behavior to create personalized rewards and incentives. In education and corporate training, we can use generative artificial intelligence to develop interactive simulations that modify task complexity following employee performance, thereby ensuring a stimulating and adaptive learning environment.

## 5 Discussion

In the educational context, the adaptive characteristics of LLMs ensure that learners are continually challenged while avoiding feeling overwhelmed. In healthcare, these models exhibit the potential to facilitate behavior modification through tailored challenges. Retail applications underscore the capacity to enhance customer loyalty by anticipating preferences and providing immediate rewards. These insights imply that the proposed framework effectively bridges the divide between user engagement and system adaptability, rendering it a significant asset across multiple fields.

The study reported in this paper contributes to modeling real-time adaptability in generative artificial Intelligence-driven gamification and establishes a framework that can be validated across various industries, in contrast to previous qualitative analyses. This work also introduces quantitative engagement and task adaptation models. The framework summarized here allows for dynamic engagement levels and task difficulty adjustments. This may increase the responsiveness of gamified systems. A brief overview of potential implementations of the framework in education, healthcare, and retail is provided, highlighting its extensive applicability.

Although this work presented here is an approach to incorporating artificial intelligence into gamified systems, it is essential to recognize certain limitations. Computational demands in processing pose significant challenges for large-scale applications, which may require the development of optimized architecture or hybrid artificial intelligence solutions. The other issues that may arise are ethical issues. The first problem may be data privacy, as personalized engagement strategies often involve substantial data collection. Second, the potential biases in content generated by generative artificial intelligence to ensure equitable and inclusive user experiences. Future studies will consider those limitations that were previously referred to. However, future studies may also focus on developing lightweight language model architectures and improving interpretability.

## 6 Conclusion

This paper suggests a mathematical framework to show how combining Large Language Models (LLMs) into gamified systems might improve user engagement, modify task dynamics, and maximize reward systems. The research offers a strong theoretical foundation for integrating LLMs into gamified systems by developing important mathematical equations and models, such as task adaptability, reward optimization, and user involvement dynamics. The simulated environment further confirmed the framework's usefulness, demonstrating how real-time modifications to user performance may produce a more immersive and customized experience.

The simulation findings show that the LLM-enhanced gamified system can maintain user interest by dynamically adjusting to each user's needs. The relationship between task difficulty and engagement showed how important it is to balance obstacles and attainable objectives to sustain motivation and interest. The framework may be customized to fit different user profiles and situations because of its flexibility in adjusting factors, including task adaption rate, reward scaling, and disengagement scaling.

The possible applications in other industries were examined, emphasizing how this integrated strategy might greatly enhance the retail, healthcare, and education sectors. For instance, LLM-powered personalized learning experiences may instantly adjust to a student's development, keeping them interested and involved. AI-powered challenges in the healthcare industry can encourage consumers to lead better lives by providing personalized guidance and engaging exercise regimens. Because of its adaptability, the framework can be used in a variety of contexts, increasing task results and user retention.

By recommending that future research concentrate on improving these models to handle more complex user behaviors, adding more machine-learning techniques, and broadening the range of applications, this paper establishes the groundwork for future studies into using LLMs in gamified environments. By making systems more responsive, engaging, and adaptive, gamification with LLMs can revolutionize businesses and offer a more efficient means of promoting user motivation and bringing about long-term behavioral change.

The upcoming work is expected to explore a real-world case study. The insights derived from this case study will be essential for comprehending the practical application of AI in educational technology and will inform future developments in adaptive learning systems.

*Acknowledgement:*
Thank you to the reviewers who gracefully gave valuable and constructive comments. The author acknowledges financial support from the Fundação para a Ciência and Tecnologia (FCT Portugal) through research grant numbers ADVANCE-CSG UIDB/04521/2020 and acknowledges the use of

Generative AI tools, such as ChatGPT, to refine the language and clarity of this paper. All intellectual contributions and research content remain his own.